\documentclass[letterpaper, 10 pt, conference]{ieeeconf}
\usepackage{amsmath}
\usepackage{amssymb} 
\usepackage{algorithm2e}
\usepackage{graphicx} 
\usepackage{array,booktabs}
\usepackage{color}
\usepackage[font = footnotesize]{caption}
\usepackage[export]{adjustbox}

\definecolor{airforceblue}{rgb}{0.36, 0.54, 0.66}
\SetAlFnt{\small}
\SetAlgoCaptionLayout{footnotesize}

\SetCommentSty{mycommfont}
\DontPrintSemicolon

\IEEEoverridecommandlockouts                              

\title{\LARGE \bf
PrimiTect: Fast Continuous Hough Voting for Primitive Detection 
}

\author{Christiane Sommer$^{1}$ \quad Yumin Sun$^{1}$ \quad Erik Bylow$^{1}$ \quad Daniel Cremers$^{1}$
\thanks{*This work was partially supported by the ERC Consolidator Grant ``3D Reloaded''.}
\thanks{$^{1}$
	All authors are with the Department of Informatics, TU Munich, Germany: \texttt{\{sommerc,suny,bylow,cremers\}@in.tum.de}}%
}

 \usepackage{tikz}
 \newcommand\copyrighttext{%
   \footnotesize \copyright 2020 IEEE. This article version was accepted for publication in IEEE International Conference on Robotics and Automation (ICRA), DOI will be provided once available. Personal use of this material is permitted.  Permission from IEEE must be obtained for all other uses, in any current or future media, including reprinting/republishing this material for advertising or promotional purposes, creating new collective works, for resale or redistribution to servers or lists, or reuse of any copyrighted component of this work in other works.}
 \newcommand\copyrightnotice{%
 \begin{tikzpicture}[remember picture,overlay]
 \node[anchor=south,yshift=5pt] at (current page.south) {\fbox{\parbox{\dimexpr\textwidth-\fboxsep-\fboxrule\relax}{\copyrighttext}}};
 \end{tikzpicture}%
 }

\begin{document}

\maketitle
\copyrightnotice
\thispagestyle{empty}
\pagestyle{empty}

\begin{abstract}
This paper tackles the problem of data abstraction in the context of 3D point sets.
Our method classifies points into different geometric primitives, such as planes and cones, leading to a compact representation of the data.
Being based on a semi-global Hough voting scheme, the method does not need initialization and is robust, accurate, and efficient.
We use a local, low-dimensional parameterization of primitives to determine type, shape and pose of the object that a point belongs to.
This makes our algorithm suitable to run on devices with low computational power, as often required in robotics applications.
The evaluation shows that our method outperforms state-of-the-art methods both in terms of accuracy and robustness.

\end{abstract}

\section{Introduction}

Many of today's imaging sensors and computer vision algorithms generate
highly accurate and dense point clouds in 3D:
monocular cameras together with (semi-)dense SLAM algorithms, deep learning techniques that predict 3D point locations, or range sensors such as RGB-D cameras or laser scanners.
With the availability of these point clouds comes the need for data abstraction and extraction of more high-level information.
A typical approach to infer such information is to train a neural network which assigns each point to a class.
This requires training data and needs GPUs for acceptable processing speed.
Applications like robotics, video surveillance or AR/VR require algorithms that can run on small embedded devices with limited computational power.

We propose a method that focuses on fast yet accurate data abstraction.
In many scenarios, a large portion of the scene can be represented by different \textit{geometric primitives}.
Our approach is based on semi-global Hough voting, which parameterizes objects locally, reducing the number of parameters and thus processing and memory requirements~\cite{drost10,drost15}.
We use linear interpolation weights to obtain continuous shape and pose estimates~\cite{niblack90}, and adapt them to optimally reflect the low-dimensional nature of 3D primitives.

\begin{figure}
	\includegraphics[width=.19\linewidth,trim={80pt 0 80pt 0},clip]{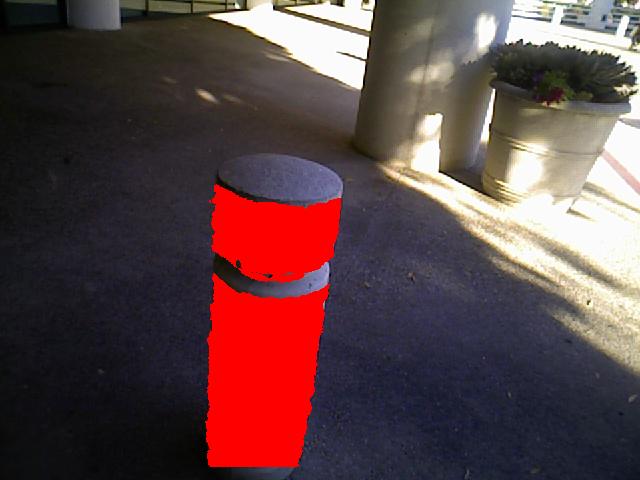}\hfill
	\includegraphics[width=.19\linewidth,trim={80pt 0 80pt 0},clip]{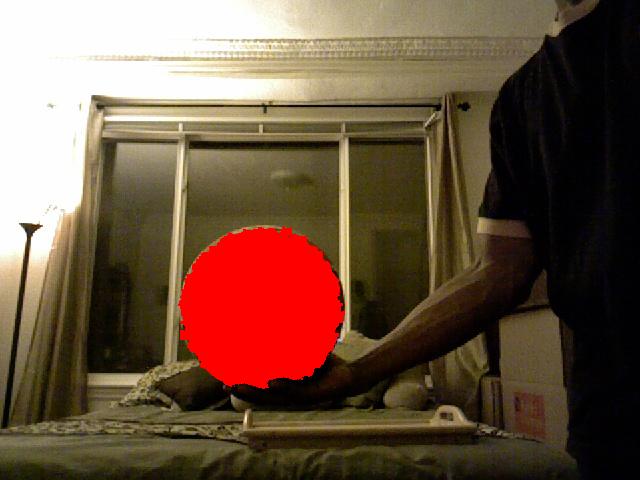}\hfill
	\includegraphics[width=.19\linewidth,trim={80pt 0 80pt 0},clip]{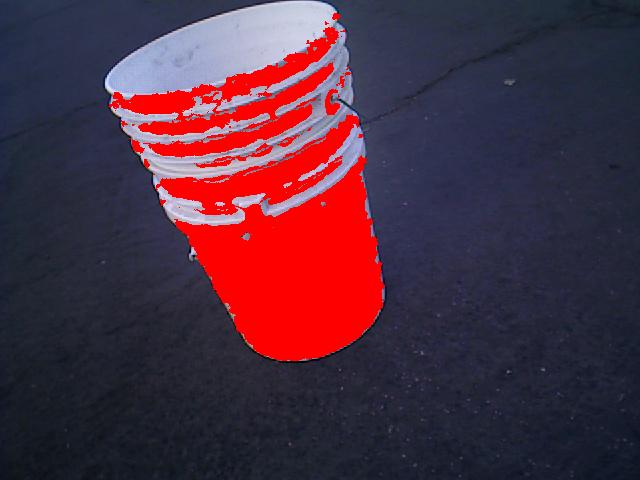}\hfill
	\includegraphics[width=.19\linewidth,trim={80pt 0 80pt 0},clip]{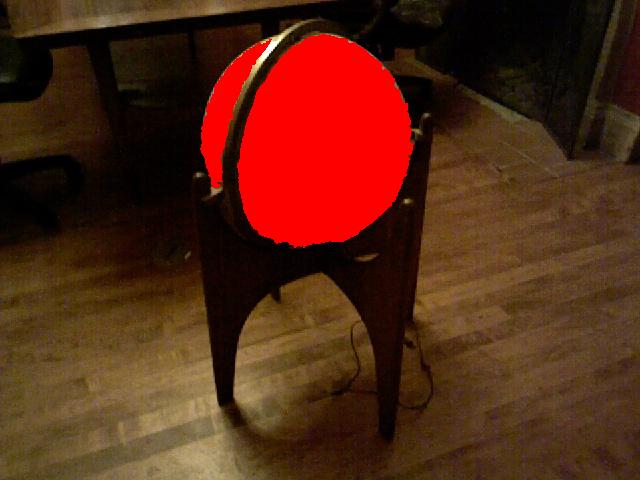}\hfill
	\includegraphics[width=.19\linewidth,trim={80pt 0 80pt 0},clip]{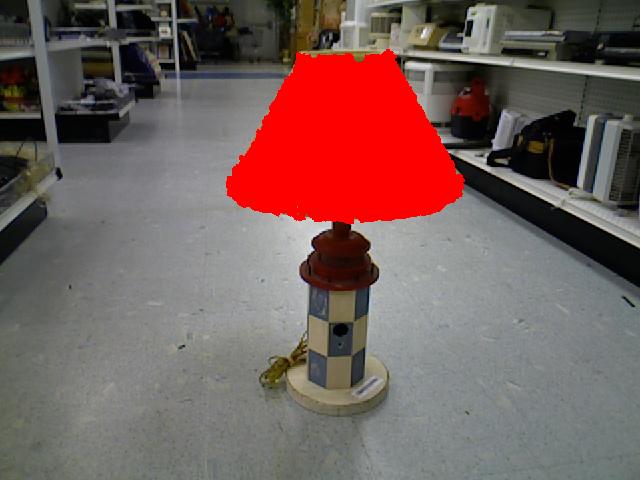}\vspace{1mm}
	\includegraphics[width=.19\linewidth,trim={80pt 0 80pt 0},clip]{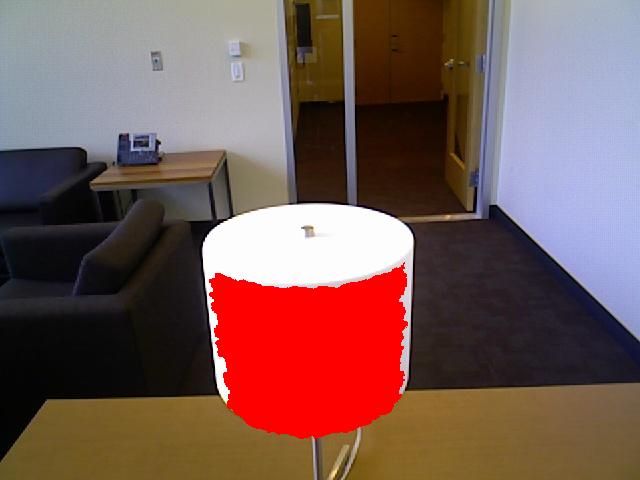}\hfill
	\includegraphics[width=.19\linewidth,trim={80pt 0 80pt 0},clip]{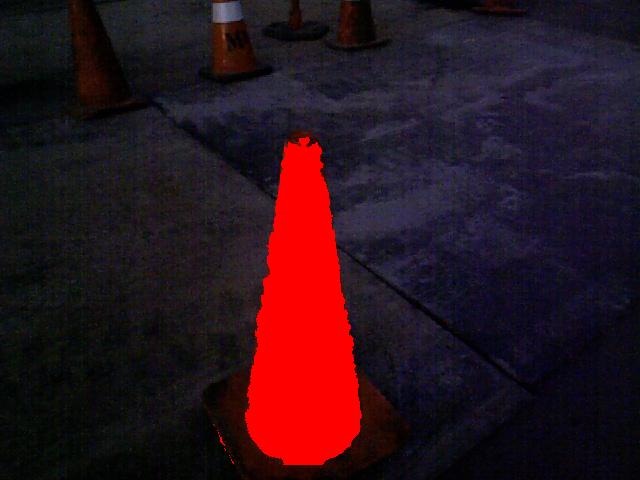}\hfill
	\includegraphics[width=.19\linewidth,trim={80pt 0 80pt 0},clip]{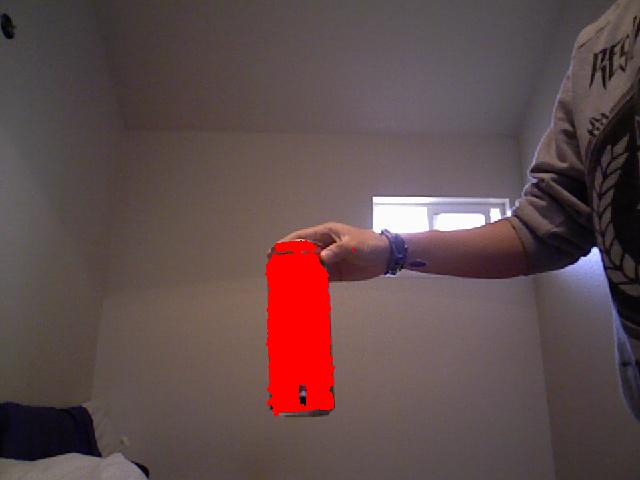}\hfill
	\includegraphics[width=.19\linewidth,trim={80pt 0 80pt 0},clip]{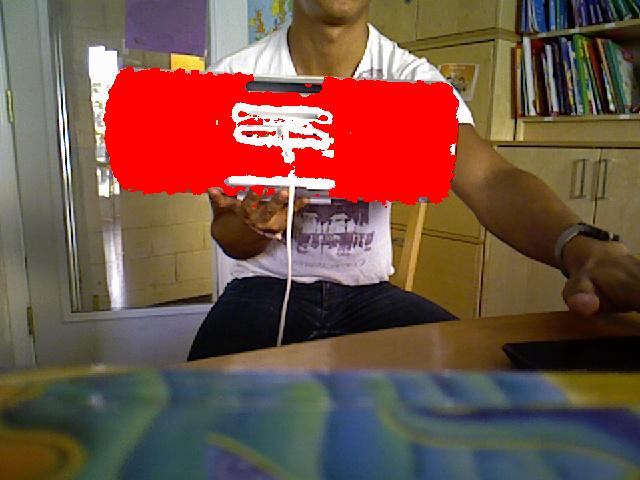}\hfill
	\includegraphics[width=.19\linewidth,trim={80pt 0 80pt 0},clip]{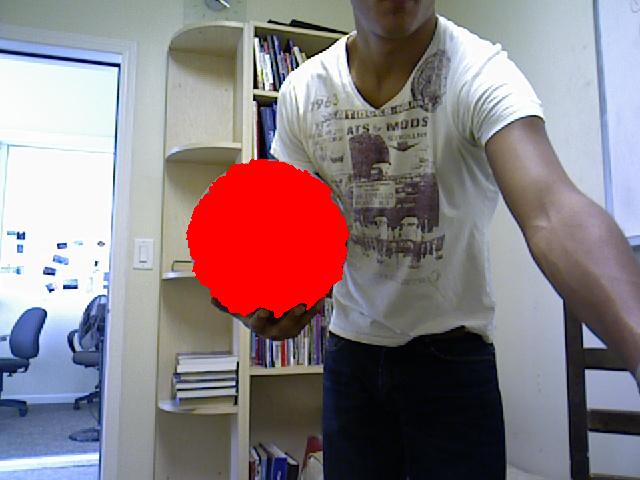}
	\caption{Geometric primitives detected on the \textit{primitive-rw}~\cite{choi16} data: our method can find spheres, cylinders, and cones with good accuracy in the depth data from a standard RGB-D camera (colors only for visualization).}
	\label{fig:redwood_results}
\end{figure}

We briefly summarize the related work, focusing on methods that can operate under clutter and occlusion. 
For a more complete overview, refer to~\cite{kaiser19}.
The most influential and de-facto standard method for primitive detection in 3D point sets is the Efficient RANSAC algorithm by Schnabel~\textit{et al.}~\cite{schnabel07}.
It randomly samples triplets of points with their associated normals, and computes the primitives that these three points define.
An inlier check validates or discards the triplet.
Several other works are inspired by RANSAC-based algorithms~\cite{li11,kang15}.
\cite{kim05}, \cite{georgiev13} and~\cite{tran15} use approaches based on region-growing.
Recently, there have also been approaches that learn to predict a set of primitives from 3D point cloud data~\cite{sharma18,li19}.
Most notably, SPFN by Li~\textit{et al.}~\cite{li19} reported promising results on a synthetic dataset of CAD models.
Their method extracts object parts from mechanical tools, but does not generalize to arbitrary types of point clouds.

The method most similar to ours is the local Hough transform proposed by Drost and Ilic~\cite{drost15}.
We adapt their idea of semi-global voting and generalize it in several ways.
We derive new voting strategies for cylinders and cones, together with ways to compute the voting parameters without model pre-creation as in~\cite{drost15}.

As our method is based on Hough voting, we also give a short overview of Hough voting, especially in the context of primitive detection.
Originally, the Hough transform was used to extract straight lines from 2D image data by casting votes into a 2D accumulator array representing the two parameters of the line~\cite{hough62,duda72}.
Several works have investigated the problem of discretization and other issues of the Hough transform~\cite{niblack90,kiryati91,zhang96,fernandes08,seib15}.
Niblack and Petkovic proposed vote spreading and weighted averaging for maximum extraction to improve the accuracy~\cite{niblack90}.

In the context of 3D primitive detection, voting-based approaches have been mostly used for plane detection, e.g.~\cite{ding05,tarshakurdi07,borrmann11,limberger15,vera18}.
Some work also deals with sphere or cylinder detection~\cite{rabbani05,su10,abuzaina13,figueiredo17}.
Birdal~\textit{et al.}~\cite{birdal18} use a voting scheme together with RANSAC initialization to detect general quadrics in point clouds.


To summarize, our main contributions in this work are:
\begin{enumerate}
	\item We derive hash-free voting conditions for cylinders and a novel voting scheme for cones using point pairs. 
	\item We generalize the concept of linear interpolation voting~\cite{niblack90} to 3D primitive detection, and extend it to constrained voting spaces, which leads to comparability of vote counts across different primitive types. 
\end{enumerate}
Furthermore, we simplify the voting decision conditions derived in~\cite{drost15} for lower computation cost. We also combine primitive detection of different types in one joint voting loop and propose a highly informative evaluation protocol.

\section{Primitive Detection Method}

The term \textit{geometric primitive} refers to different things depending on context.
In this work, we consider planes, spheres, cylinders and cones as geometric primitives.
We assume as input a point cloud $\mathcal{P}=\left\{\mathbf{p}_j\right\}_j$ together with oriented normals $\{\mathbf{n}_j\}_j$.

\subsection{Point pair features and voting}
\label{sec:ppf_definitions}

Given two points $\mathbf{p}_r$, $\mathbf{p}_i$ with corresponding normals $\mathbf{n}_r$, $\mathbf{n}_i$, the point pair feature (PPF) vector $F(\mathbf{p}_r,\mathbf{p}_i)\in\mathbb{R}_{\geq 0}\times (0,\pi)^3$ is defined as~\cite{drost10}
\begin{equation}
F(\mathbf{p}_r,\mathbf{p}_i) = (\Vert\mathbf{d}\Vert, \angle(\mathbf{d},\mathbf{n}_r), \angle(\mathbf{d},\mathbf{n}_i), \angle(\mathbf{n}_r,\mathbf{n}_i))\:,
\end{equation}
where $\mathbf{d} = \mathbf{p}_i-\mathbf{p}_r$.
We define a new 4D feature vector $C$ that is strongly related to $F$ by
\begin{equation}
\begin{aligned}
C(\mathbf{p}_r, \mathbf{p}_i) &=
\left(\Vert\mathbf{d}\Vert^2, \mathbf{n}_r^\top\mathbf{d}, \mathbf{n}_i^\top\mathbf{d}, \mathbf{n}_r^\top\mathbf{n}_i\right) = \\
&= (F_1^2, F_1\cos(F_2), F_1\cos(F_3), \cos(F_4))\:.
\end{aligned}
\end{equation}
The advantage of $C$ over $F$ is that it can be computed without expensive trigonometric operations 
while the information content is the same.

Given a point cloud $\mathcal{P}$, in PPF-based approaches such as~\cite{drost15}, a set of reference points $\{\mathbf{p}_r\}_r$ is selected and the following steps are executed for each of them~\cite{drost10}:
\begin{enumerate}
	\item Initialize the voting accumulator array.
	\item Pair $\mathbf{p}_r$ with a subset $\{\mathbf{p}_i\}_i$ of $\mathcal{P}$ and compute the corresponding point pair features $C(\mathbf{p}_r,\mathbf{p}_i)$.
	\item For each point pair $\mathbf{p}_r$, $\mathbf{p}_i$, vote for the object/pose that it could belong to.
	\item Extract the object/pose candidate with the maximal number of votes from the accumulator array.
\end{enumerate}
In the end, the extracted candidates are clustered.
\cite{drost15}~used a simple approach for clustering based on non-maximum suppression (NMS), while we compute a weighted average of similar candidates, with the weights being the vote count.

\begin{table}
	\caption{Voting space dimensions~\cite{drost15}, voting parameters and number of equality constraints for the four different types of primitives.}
	\label{table:voting_dims}
	\begin{center}
		\renewcommand{\arraystretch}{1.25}
		\begin{tabular}{l  c  c  c  c}
			\toprule
			& plane & sphere & cylinder & cone \\
			\midrule
			global dim. of pose & 3 & 3 & 4 & 5 \\
			local dim. of pose & 0 & 0 & 1 & 2 \\
			\# shape params & 0 & 1 & 1 & 1 \\
			local dim. in total & 0 & 1 & 2 & 3 \\
			voting parameters & - & $R$ & $(R,\varphi)$ & $(s_r,\mathbf{a})$ \\
			voting space & $\{c\}$ & $\mathbb{R}^{+}$ & $\mathbb{R}^{+}\times[0,\pi)$ & $\mathbb{R}^{+}\times\mathbb{S}^2$ \\
			\# equality constraints & 3 & 2 & 1 & 0 \\
			\bottomrule
		\end{tabular}
	\end{center}
\end{table}

\subsection{Voting conditions for geometric primitives}
\label{sec:voting_conditions}

Generic rigid objects can be detected by hashing the PPFs of a model point cloud~\cite{drost10}.
For geometric primitives we can derive conditions on $C(\mathbf{p}_r,\mathbf{p}_i)$ (or, equivalently, on $F$) that must hold true.
As all four primitives are convex, we have the inequality constraints $C_2\leq0$ and $C_3\geq0$.
Furthermore, we have the following conditions:

\paragraph{Planes}
Two points exactly lying on a common plane have equality constraints on three components of $C$:
%
\begin{equation}
C(\mathbf{p}_r,\mathbf{p}_i) = \left(\Vert\mathbf{d}\Vert^2, 0, 0, 1\right)\:.
\end{equation}
$(\mathbf{p}_r,\mathbf{n}_r)$ already fully define the plane, thus the local plane voting space is zero-dimensional.

\begin{figure}
	\includegraphics[width=.33\linewidth]{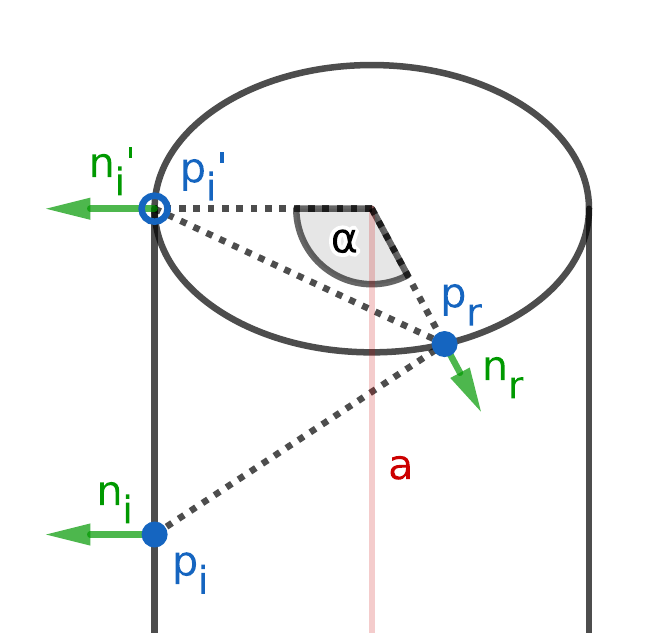}\hfill
	\includegraphics[width=.33\linewidth]{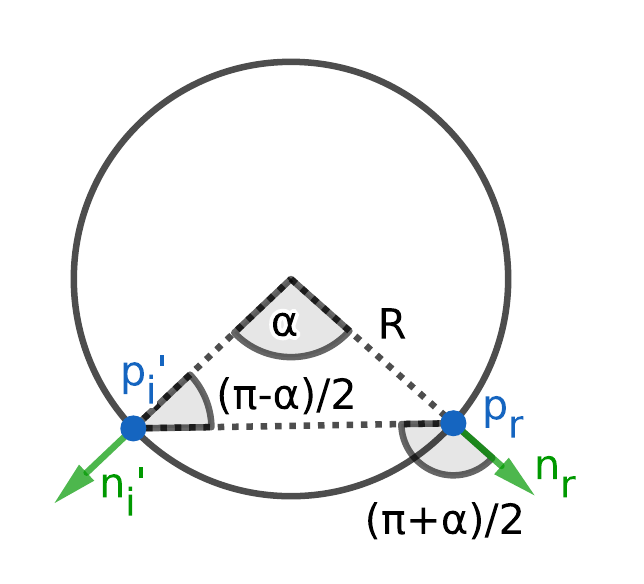}\hfill
	\includegraphics[width=.33\linewidth]{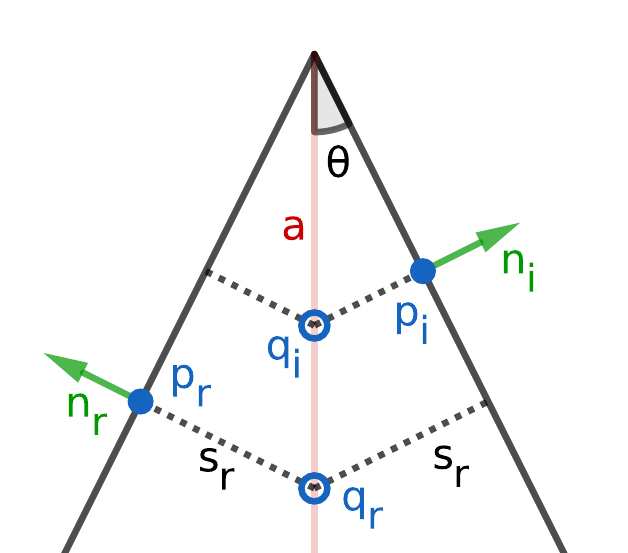}
	\caption{Angles and distances involved in PPFs and voting parameters. \textbf{Left:} for the cylinder, we shift $\mathbf{p}_i$ along the axis up to the height of $\mathbf{p}_r$ to obtain $\mathbf{p}_i'$. For the sphere, we set $\mathbf{p}_i'=\mathbf{p}_i$. \textbf{Middle:} As $\mathbf{p}_i'-\mathbf{p}_r$, $\mathbf{n}_r$ and $\mathbf{n}_i$ are co-planar, we illustrate the angle relations for sphere and cylinder in 2D. \textbf{Right:} We obtain points on the cone axis by going from $\mathbf{p}_r$ and $\mathbf{p}_i$ in their negative normal directions: $\mathbf{a}\parallel \mathbf{q}_i-\mathbf{q}_r$.}
	\label{fig:ppf_angles}	
\end{figure}

\paragraph{Spheres}
For a sphere with radius $R$, the PPF is given as (see also Fig.~\ref{fig:ppf_angles}, middle)~\cite{drost15}
\begin{equation}
\label{eq:sphere_ppf}
F(\mathbf{p}_r,\mathbf{p}_i) = \left(2R\sin\left(\frac{\alpha}{2}\right),\frac{\pi+\alpha}{2},\frac{\pi-\alpha}{2},\alpha\right)
\end{equation}
with $\alpha=\angle(\mathbf{n}_r,\mathbf{n}_i)$.
In particular, we have the two constraints $F_2+F_3=\pi$ and $F_2-F_3=F_4$.
We apply angle-sum and half-angle identities on Eq.~\eqref{eq:sphere_ppf} to obtain
\begin{equation}
C(\mathbf{p}_r,\mathbf{p}_i) = \left(2\lambda R^2,-\lambda R, \lambda R, 1-\lambda\right)
\end{equation}
with $\lambda=1-{\mathbf{n}_r}^\top\mathbf{n}_i$. Now, we can easily compute the radius
\begin{equation}
\label{eq:radius_sphere}
R = \frac{C_2-C_3}{2(C_4-1)}\:,
\end{equation}
which, together with $(\mathbf{p}_r,\mathbf{n}_r)$, uniquely defines the sphere.
Note that our formula for the radius is different from and less computationally expensive than the one given in~\cite{drost15}.

\paragraph{Cylinders}
For a cylinder, we first make the important observation that if we shift $\mathbf{p}_i$ along the cylinder axis $\mathbf{a}\in\mathbb{S}^2$ such that it is at the same height as $\mathbf{p}_r$ (see Fig.~\ref{fig:ppf_angles} for geometric illustration),
\begin{equation}
\mathbf{p}_i'=\mathbf{p}_i + \mathbf{a}^\top(\mathbf{p}_r-\mathbf{p}_i)\cdot\mathbf{a}\:,
\end{equation}
we can use $\mathbf{a}^\top\mathbf{n}_r=\mathbf{a}^\top\mathbf{n}_i=0$ to obtain
\begin{equation}
C_2 = (\mathbf{p}_i'-\mathbf{p}_r)^\top\mathbf{n}_r\:,\quad C_3 = (\mathbf{p}_i'-\mathbf{p}_r)^\top\mathbf{n}_i\:.
\end{equation}
Thus, the constraint $F_2+F_3=\pi$ as well as $C_2$, $C_3$ and $C_4$ can be derived in exact analogy to the sphere PPF:
\begin{equation}
C(\mathbf{p}_r,\mathbf{p}_i) = \left(\Vert\mathbf{d}\Vert^2,-\lambda R,\lambda R, 1-\lambda\right)\:.
\end{equation}
%
%
In particular, we can re-use Eq.~\eqref{eq:radius_sphere} for radius computation.
Thus, although it is claimed in~\cite{drost15} that an implicit model for cylinder PPFs is quite complex, we provide a very simple way to compute the radius without any model pre-creation.
There is one local (voting) parameter for the pose of the cylinder -- the angle $\varphi$ of the rotation axis in a local coordinate system on the tangent plane at $\mathbf{p}_r$.
We compute $\varphi$ similarly to~\cite{drost10}:
we align $\mathbf{n}_r$ with the positive $x$-axis by a rotation $R_x$
and define $\varphi$ as the angle with the $y$-axis in the rotated coordinate system.
Note that $\varphi$ needs to be computed even if a cylinder model is pre-created.
%
The cylinder axis $\mathbf{a}$ can be retrieved from $\varphi$ and $R_x$ as $((0,\cos\varphi,\sin\varphi)R_x)^\top$.

\paragraph{Cones}
Cones are not handled in~\cite{drost15}.
Thus, we derive a novel PPF-based voting scheme for cones.
As we illustrate in Fig.~\ref{fig:ppf_angles}, a point $\mathbf{q}$ on the cone axis $\mathbf{a}$ has the same distance to any tangent plane, in particular
\begin{equation}
\label{eq:tangent_dist}
\left(\mathbf{q}-\mathbf{p}_i\right)^\top\mathbf{n}_i = \left(\mathbf{q}-\mathbf{p}_r\right)^\top\mathbf{n}_r\:.
\end{equation}
We define $s_r$ such that $\mathbf{q}_r=\mathbf{p}_r-s_r\mathbf{n}_r$ is an axis point and get
\begin{equation}
\label{eq:compute_sr}
-s_r = \left(\mathbf{q}_r-\mathbf{p}_i\right)^\top\mathbf{n}_i = -C_3 - s_rC_4\:.
\end{equation}
The first equality follows from Eq.~\eqref{eq:tangent_dist}, while the second one is a consequence of the definition of $\mathbf{q}_r$.
An analogous definition of $\mathbf{q}_i$, and corresponding derivations allow us to compute the cone axis direction $\mathbf{a}$, which is parallel to
\begin{equation}
\mathbf{q}_i-\mathbf{q}_r = \mathbf{d} - \frac{C_3\mathbf{n}_r+C_2\mathbf{n}_i}{C_4-1}\:.
\end{equation}
Note that $\mathbf{p}_r$, $\mathbf{p}_i$ do not need to satisfy any equality constraint to lie on a cone.
We vote for $s_r$, which can be computed from Eq.~\eqref{eq:compute_sr}, and the axis $\mathbf{a}\in\mathbb{S}^2$
using an accumulator ball~\cite{borrmann11}.
The accumulator ball yields nearly homogeneous bin sizes, leading to less voting bias.
As the axis direction is only unique up to sign at this stage, we create only a hemisphere accumulator.

Upon candidate extraction, the cone parameters (apex $\mathbf{c}$, signed axis $\mathbf{a}$, and opening angle $\theta$)  can be derived from the maximum vote parameters $(\hat{s}_r,\hat{\mathbf{a}})$ and the reference point:
\begin{align}
\mathbf{c} &= \mathbf{p}_r + \hat{s}_r\left(\frac{\hat{\mathbf{a}}}{\hat{\mathbf{a}}^\top\mathbf{n}_r}-\mathbf{n}_r\right)\:,\\
\mathbf{a} &= \operatorname{sgn}((\mathbf{p}_r-\mathbf{c})^\top\hat{\mathbf{a}})\cdot\hat{\mathbf{a}}\:,\\
\sin\theta &= -\mathbf{a}^\top\mathbf{n}_r\:.
\end{align}

\begin{table*}
	\caption{Voting conditions: from exact conditions on $F(\mathbf{p}_1,\mathbf{p}_2)$ to relaxed conditions on $C(\mathbf{p}_1,\mathbf{p}_2)$.}
	\label{table:voting_conditions}
	\begin{center}
		\renewcommand{\arraystretch}{1.25}
		\begin{tabular}{c | c | c | c }
			\toprule
			& exact ($F$) & relaxed ($F$) & relaxed ($C$) \\
			\midrule
			normals parallel (NP) & $F_4 = 0$ & $|F_4|<\epsilon_\textsc{np}$ & $C_4>\cos\epsilon_\textsc{np}$ \\
			points co-planar (PC) & $F_2=F_3=\tfrac{\pi}{2}$ & $|F_{2,3}-\tfrac{\pi}{2}|<\epsilon_\textsc{pc}$  & $C_2^2 < (\sin\epsilon_\textsc{pc})^2 C_1$ and $C_3^2 < (\sin\epsilon_\textsc{pc})^2 C_1$ \\
			angles symmetric (AS) & $F_2+F_3 = \pi$ & $|F_2+F_3-\pi| < \epsilon_\textsc{as}$ & $S_2S_3-C_2C_3 > \cos\epsilon_\textsc{as}C_1$ with $S_2 = \sqrt{C_1-C_2^2}$, $S_3 = \sqrt{C_1-C_3^2}$ \\
			vectors triangular (VT) & $F_2-F_3=F_4$ & $|F_2-F_3-F_4| < \epsilon_\textsc{vt}$ & $C_2C_3C_4 + S_2S_3C_4 + S_2C_3S_4 - C_2S_3S_4 > \cos\epsilon_\textsc{vt}C_1$, $S_4=\sqrt{1-C_4^2}$\\
			\bottomrule
		\end{tabular}
	\end{center}
\end{table*}

We summarize all voting parameters, voting space dimensions and number of equality constraints in Table~\ref{table:voting_dims}.
In real world data, neither point positions nor the computed normals are exact, which results in non-exact PPFs.
Therefore, all the derived constraints need to be relaxed.
Using angle addition theorems, we rewrite the relaxed conditions in terms of $C(\mathbf{p}_r, \mathbf{p}_i)$, eliminating the need for trigonometric operations.
The final conditions are summarized in Table~\ref{table:voting_conditions}.

\subsection{Joint voting}
\label{sec:joint_voting}

While~\cite{drost15} propose detectors for different primitive types, there is no joint detector, such as~\cite{schnabel07}.
We suggest a joint voting process, which is outlined in Algorithm~\ref{alg:voting_decisions}.
This allows for multi-primitive detection in scenes that contain more than one primitive type.

\begin{algorithm}[t]
	\SetAlgoLined
	\KwData{point pair feature $C\in \mathbb{R}_{\geq 0}\times[-1,1]^3$,\qquad accumulator arrays for all four primitives}
	\KwResult{updated accumulator arrays}
	\If{$C_2>0$ or $C_3<0$}{
		return
	}
	\eIf{normals parallel (NP)}{
		\If{points co-planar (PC)}{
			vote for plane \tcp*[1]{3 constraints, 0D voting space}
		}
	}{
	compute $s_r$ and cone axis $\mathbf{a}$ \\
	vote for cone $(s_r,\mathbf{a})$  \tcp*[1]{0 constraints, 3D voting space}
	\If{angles symmetric (AS)}{
		compute radius $R$ and cylinder angle $\varphi$\\
		vote for cylinder $(R,\varphi)$ \tcp*[1]{1 constraint, 2D v. s.}
		\If{vectors form triangle (VT)}{
			vote for sphere $(R)$ \tcp*[1]{2 constraints, 1D v. s.}
		}
	}
}
\caption{Voting decisions for joint primitive detection: plane PPFs are disjoint from all other types. Sphere PPFs are a subset of cylinder PPFs, which are a subset of cone PPFs. This is reflected in the code structure. The conditions NP, PC, AS, and VT are defined in Table~\ref{table:voting_conditions}.}
\label{alg:voting_decisions}
\end{algorithm}

In cases where cylinders and cones have parallel normals, the voting parameters cannot be determined without ambiguity.
For the sphere, parallel normals cannot even arise.
Thus, we do not vote for sphere, cylinder or cone in these cases. 
Effectively, this makes the plane case disjoint from the other three objects, which significantly reduces computation, as usually scenes exhibit large numbers of planar structures.

\subsection{Linear interpolation voting}
\label{sec:linear_interpolation}

For the classical application of Hough transforms, Niblack and Petkovic proposed to spread votes over different bins to overcome the limitation arising from the discrete nature of voting bins~\cite{niblack90}. 
After voting, parameters are extracted as a weighted average of the parameters in the neighborhood of the bin which has the most votes.
In an ideal noise-free world, \textit{linear interpolation weights} can perfectly reproduce the original parameters. 
%
Pulling this concept into the world of 3D primitive detection poses several challenges:
\begin{enumerate}
	\item It is not clear a priori how the concept of linear interpolation voting translates to the constraints (see Table~\ref{table:voting_conditions}) that effectively reduce the voting space dimensionality.
	\item $d$-dimensional linear interpolation weights are on average $2^{-d}$, so that while a sphere vote (1D voting space) is on average ${1}/{2}$, a cone vote (3D) will be only ${1}/{8}$.
	We need to find a way to make the vote counts comparable across objects.
	\item The voting accumulator for the cone axis is a discrete version of $\mathbb{S}^2$, so we need a notion of linear interpolation weights on the unit sphere that is not too computationally expensive.
\end{enumerate}
First, we notice that the number of constraints and the dimensionality of the voting space add up to $3$ for all four types of primitives.
Thus, for each constraint, we add a weight similar to the linear interpolation weights that measures how closely the exact condition is met.
E.g., for the cylinder, we need to fulfill AS, and thus we multiply each vote by a \textit{constraint weight}
\begin{equation}
w_{\textsc{as}} = \left(1 - \frac{|F_2 + F_3 - \pi|}{\epsilon_\textsc{as}}\right)\:.
\end{equation}
Introducing equivalent weights for the other constraints, we successfully resolve issue 1).
Choosing the $\epsilon$ according to the angular voting bin size leads to the desired comparability mentioned in 2).
Lastly, we resolve 3) by using a flat approximation of $\mathbb{S}^2$ parameterized by $x$ and $y$ in the pole neighborhoods, and a parameterization by $\vartheta$ and $\varphi$ elsewhere.
While this leads to non-exact linear interpolation weights, it is a trade-off between speed and accuracy of weights.

\section{Experimental Setup}

\subsection{Datasets}

We run our algorithm both on synthetic and real-world datasets of different types.

The authors of SPFN~\cite{li19} released the \textit{traceparts} dataset on which their method is trained.
It consists of 18k point clouds of mechanical parts, which were obtained by converting CAD models to point clouds and adding uniform noise.
Some of the shapes contain concave cylinders, i.e. cylinders for which the normals are pointing inwards (e.g. inner walls of a tube).
Our method assumes convexity of primitives, so we cannot fairly evaluate on \textit{traceparts} and only provide a very basic evaluation of this.


We use a subset of the real-world \textit{redwood} dataset~\cite{choi16} which consists of a large number of RGB-D images without groundtruth.
Images were selected that contained a primitive in the foreground and the depth was truncated at 2m following~\cite{hodan18}.
In total, about 100 point clouds were selected.
We call this modified dataset \textit{primitive-rw}.


To evaluate the performance for different primitive types separately, we need data that only contains one specific object type.
However, we also want to evaluate the joint detection on the same type of data. 
Thus, we introduce our own \textit{primitect} dataset by rendering a number of primitives (max. 20) in a stereographic projection manner and add Gaussian noise.
\textit{primitect} features a total of 175 point clouds with 160k points each.
We focus on spheres, cylinders, cones and joint detections.


\subsection{Evaluation metrics}
\label{sec:evaluation}

While~\cite{li19} propose a whole set of evaluation metrics, we find plots to be more intuitive to understand than tables.
We introduce two types of cumulative plots. 

\paragraph{p-coverage plot}
we use the \textit{p-coverage} from~\cite{li19}:
for each point $\mathbf{p}_i$ in the point cloud, we compute its (positive) distance $e_i$ to its closest object,
\begin{equation}
e_i = \min_k{d(\mathbf{p}_i, \hat{O}_k)}\:,
\label{eq:dist_func}
\end{equation}
where $\{\hat{O}_k\}_k$ is the set of detected primitives and $d(\mathbf{p},O)$ the distance of a point $\mathbf{p}$ to a primitive $O$.
The \textit{p-coverage} at $\epsilon$, denoted by $p(\epsilon)$, is defined as the percentage of points for which $e_i<\epsilon$. 
In~\cite{li19}, $p(\epsilon)$ is reported for $\epsilon=0.01$ and $0.02$.
We plot $p$ as a function of $\epsilon$ to get a more complete picture.
One noteworthy feature of this plot is that the area between the $p=1$ line and the curve is equal to the mean $\mu$ of the $\{e_i\}_i$~\cite{bi03}. 

\paragraph{s-coverage plot}
We do the same thing for each detected primitive $\hat{O}_k$ for all of its inliers, and call the result $p_k(\epsilon)$.
We then plot the average over all detected primitives,
and dub this function \textit{s-coverage} to emphasize its similarity to the $\{S_k\}$-coverage in~\cite{li19}.
Note, that our \textit{s-coverage} uses distances of actual data points to primitives.
We make this choice to generalize the metric to cases where no ground truth is available.

\paragraph{data-aware object distance}
For the actual primitive parameters (axes, radii, etc.), it is hard to find an error metric that is consistent across different primitive types and parameterizations:
primitive fits which are visually of very different quality can translate to the same error in parameter space, and vice versa.
We propose a more data-aware metric to judge the quality of primitive parameters:
for a given point set $\{\mathbf{p}_i\}_i$ of cardinality $N$, a ground truth primitive $O$ and a detected instance $\hat{O}$, we compute the \textit{data-aware object distance (DOD)} $D(O, \hat{O})$ by projecting each data point to $O$ and $\hat{O}$, and computing the average distance:
\begin{equation}
D(O, \hat{O}) = \frac{1}{N}\sum_{i=1}^{N}{\Vert\textrm{proj}(\mathbf{p}_i, O)-\textrm{proj}(\mathbf{p}_i, \hat{O})\Vert}\:.
\end{equation}
This choice of object distance has several advantages:
1)~It can be computed for generic object types, not only for those whose parameter space contains $\mathbb{S}^2$ as a subspace (thus providing a way to compare primitive detection to generic object detection).
2)~It is invariant to changes of coordinate systems (which is also true for an axis angle error as in~\cite{li19}, but not for general parameter-space errors).
3)~\textit{DOD} can be seen as an adaptation of the \textit{visible surface discrepancy (VSD)}~\cite{hodan18} to general point cloud data. \textit{VSD} is the current standard for evaluating object detection and pose estimation in RGB-D images.
Our definition of \textit{DOD} is similar, but not equivalent to the \textit{mean $\{S_k\}$ residual} of~\cite{li19}.

\paragraph{precision/recall}
In addition to accuracy evaluation metrics, we also evaluate the detection rate.
For this, we compute \textit{precision}, \textit{recall}, \textit{missed-rate} and \textit{noise-rate}, normalizing the metrics from SegComp~\cite{hoover96} (with $T=0.6$) by the number of detected and ground truth objects, respectively:
\begin{align}
\textit{precision} &= {(\textit{correct detections})}/{\textit{detections}}\:, \\
\textit{recall} &= {(\textit{correct detections})}/{(\textit{GT objects})}\:, \\
\textit{missed-rate} &= {\textit{missed}}/{(\textit{GT objects})}\:, \\
\textit{noise-rate} &= {\textit{noise}}/{\textit{detections}}\:.
\end{align}
Note that due to possible instances of over- and under-segmentation, \textit{precision} and \textit{noise-rate} do not necessarily sum up to $1$, and neither do \textit{recall} and \textit{missed-rate}.

\subsection{Choice of parameters}

For each point cloud, we randomly choose $2048$ reference points and pair them with $2048$ points each.
Our angle bin size is $10^\circ$, and the radius/$s_r$ bin size is $0.005d_s$, where $d_s$ is the scene diameter.
We have 40 radius/$s_r$ bins, i.e. the maximal parameter is $0.2d_s$.
A vote is cast for a point pair if $\Vert\mathbf{d}\Vert\leq 0.2d_s$.
The object/pose with maximum vote count is extracted as candidate if it has more than eight votes.
Each primitive candidate is stored together with its reference point $\mathbf{p}_r$, and two primitives $O_i$, $O_j$ are clustered if they are of the same type and 
\begin{equation}
\vert d(\mathbf{p}_i, O_j)\vert < .01d_s\:,\quad \angle\left({\mathbf{n}_i},\nabla d(\mathbf{p}_i, O_j)\right) < 20^\circ\:.
\end{equation}

\section{Results}

\begin{table*}
\caption{Precision, recall, missed-rate and noise-rate (all given in percent) for Efficient RANSAC (a), the baseline (b) and our method.
	For the data-aware object distance DOD, we also report ablation results on our method without clustering-by-averaging (c), without bin-averaging (d) and without spreading the votes (e).
	Evaluation was done on the \textit{primitect} dataset with noise $\sigma=0.01$, and the DOD is reported in multiples of $\sigma$.
	}
\label{table:segcomp_metrics}
\begin{center}
	\renewcommand{\arraystretch}{1.25}
	\begin{tabular}{c | c c c | c c c | c c c | c c c | c c c c c c}
		\toprule
		& \multicolumn{3}{c}{\textbf{precision}} & \multicolumn{3}{c}{\textbf{recall}} & \multicolumn{3}{c}{\textbf{missed}} & \multicolumn{3}{c}{\textbf{noise}} & \multicolumn{6}{c}{\textbf{data-aware object distance}} \\
		& (a) & (b) & ours & (a) & (b) & ours & (a) & (b) & ours & (a) & (b) & ours & (a) & (b) & (c) & (d) & (e) & ours \\
		\midrule
		spheres & 81.0 & 85.1 & \textbf{94.9} & \textbf{98.6} & 95.3 & 91.3 & \textbf{1.4} & 2.9 & 8.5 & 18.3 & 0.8 & \textbf{0.3} &
		1.01 & 1.32 & 1.15 & 0.49 & 0.58 & \textbf{0.48} \\
		cylinders & 54.9 & 67.4 & \textbf{95.9} & 69.8 & 65.1 & \textbf{76.4} & \textbf{13.7} & 23.9 & 23.3 & 9.0 & 6.1 & \textbf{0.7} &
		1.45 & 1.64 & 1.31 & 0.65 & 0.63 & \textbf{0.60} \\
		cones & 41.2 & 64.3 & \textbf{91.4} & \textbf{63.0} & 46.5 & 54.1 & \textbf{21.5} & 45.8 & 44.3 & 30.8 & 9.0 & \textbf{1.9} &
		1.35 & 1.54 & 1.24 & 0.95 & 0.75 & \textbf{0.70} \\
		mixed & 61.0 & 63.8 & \textbf{74.7} & \textbf{76.7} & 57.7 & 76.6 & \textbf{9.9} & 36.4 & 21.7 & 7.6 & 7.3 & \textbf{1.4} &
		1.13 & 1.39 & 1.14 & 0.59 & 0.62 & \textbf{0.56} \\
		\bottomrule
	\end{tabular}	
\end{center}
\end{table*}

\paragraph{Comparison to other methods}
Our method is compared against Efficient RANSAC~\cite{schnabel07}, the state-of-the-art method for years, using the implementation in CGAL~\cite{oesau19} with default adaptive parameters.
The detection pipeline of Drost and Ilic~\cite{drost15} is also implemented as a baseline method, with the difference that we use our radius computation for the cylinder and added the cone voting.
We refer to this implementation as \textit{Drost*}.
Since we want to evaluate the detection part only, we do not use post-refinement for any method.
A comparison against SPFN~\cite{li19} is not entirely fair, as SPFN is designed to also detect concave primitives, while it has problems detecting anything at all in point clouds that do not represent mechanical parts.
By contrast, our algorithm is a generic method for primitive detection, but cannot detect concave shapes due to our convexity constraints.

\paragraph{Ablation study}
We show in what way our different contributions improve upon~\cite{drost15} in Table~\ref{table:segcomp_metrics}.
For this, we look at five different versions of our algorithm:
\begin{itemize}
	\item the baseline version, which is the same as~\cite{drost15} for planes and spheres ((b) in Table~\ref{table:segcomp_metrics}).
	\item the full version of our algorithm (``ours''). 
	\item three versions where we switch off one of our introduced features, respectively:
	\begin{itemize}
		\item with NMS clustering as described in~\cite{drost15}, instead of clustering-by-averaging (c),
		\item without weighted averaging of neighboring bin centers (d),
		\item without linear interpolation vote spreading and constraint weights (e).
	\end{itemize}
\end{itemize}

\subsection{Detection performance}

To measure the detection performance of our proposed algorithm, we report precision, recall, missed-rate and noise-rate on the \textit{primitect} dataset with noise $\sigma=0.01$ in Table~\ref{table:segcomp_metrics}.
We find that our method is significantly better than the baseline method~\cite{drost15}.
In terms of \textit{recall} and \textit{missed-rate}, Efficient RANSAC~\cite{schnabel07} outperforms our algorithm, while we are clearly better in \textit{precision} and \textit{noise-rate}.
Hence, RANSAC is better in detecting as many instances as possible, while accepting more false positives.
Our proposed method performs particularly well for sphere and cylinder detection, while the cone detection suffers from a very high missed-rate.
This might be due to the non-linear nature of the accumulator array for the cone axis, together with the relatively high dimensionality of the voting space. 

\begin{figure}
	\includegraphics[width=.47\linewidth]{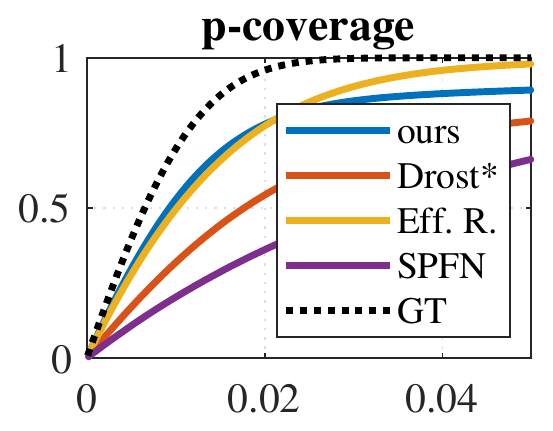}
	\hfill
	\includegraphics[width=.47\linewidth]{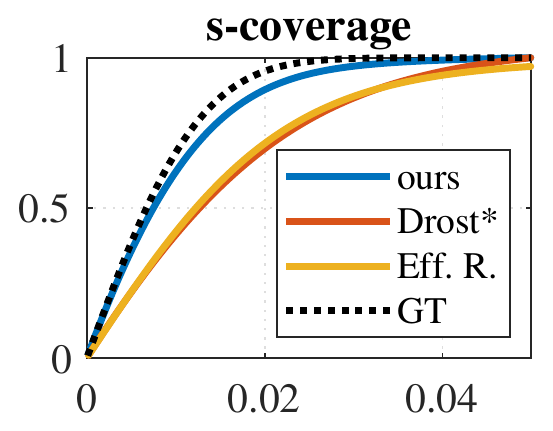}
	\caption{\textbf{(left)} p-coverage on the \textit{primitect} dataset (mixed types): Our method is better than/on par with Efficient RANSAC for $\epsilon\leq0.02$, for larger distances, the lower recall leads to lower percentages.
		\textbf{(right)} s-coverage on the same data:
		our method  fits the data much better than either the baseline and Efficient RANSAC.}
	\label{fig:coverages}
\end{figure}

A second metric that implicitly evaluates detection rate is \textit{p-coverage}. It measures the percentage of points within a certain distance to their closest geometric primitive.
The more primitives are detected, the higher this percentage is.
Fig.~\ref{fig:coverages} shows that, as expected from Table~\ref{table:segcomp_metrics}, our method stagnates, while RANSAC nearly reaches 100\%.
However, we also find that up to $2\sigma$, we are slightly better that RANSAC, meaning that our estimated parameters fit the data better when the primitive is detected.

\begin{figure}
	\includegraphics[width=.3\linewidth]{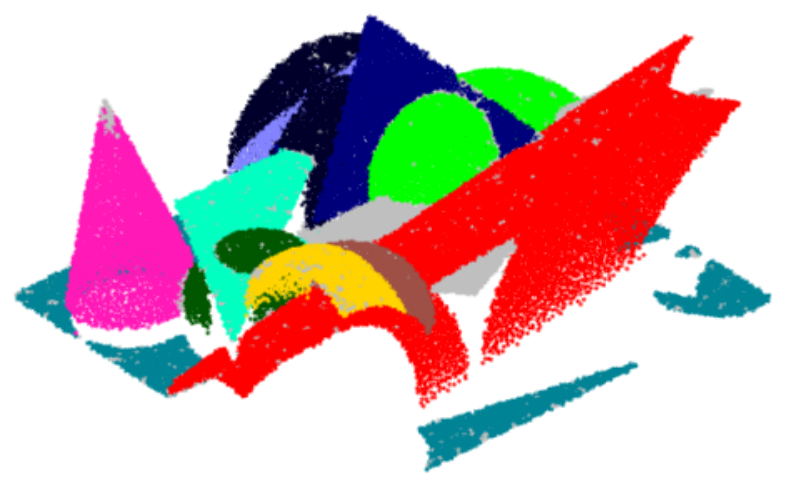}\hfill
	\includegraphics[width=.3\linewidth]{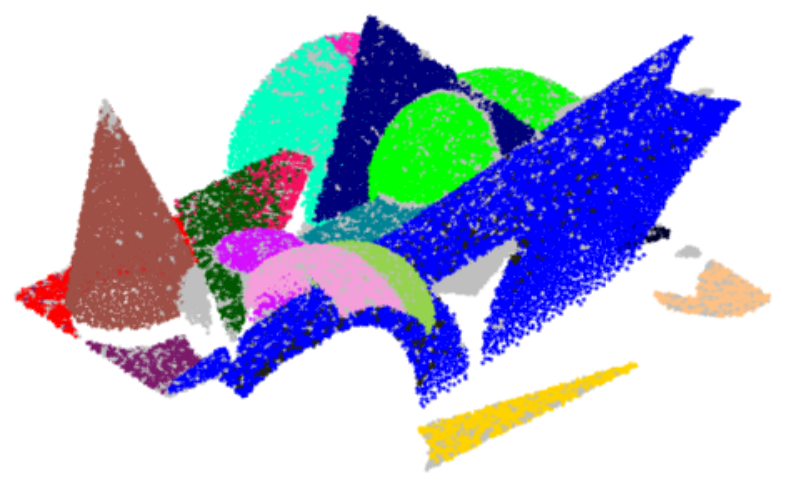}\hfill
	\includegraphics[width=.3\linewidth]{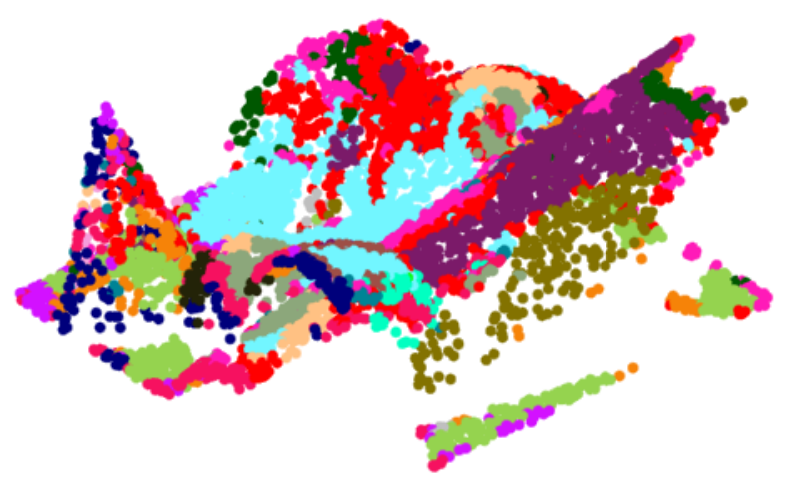}\\
	\includegraphics[width=.3\linewidth]{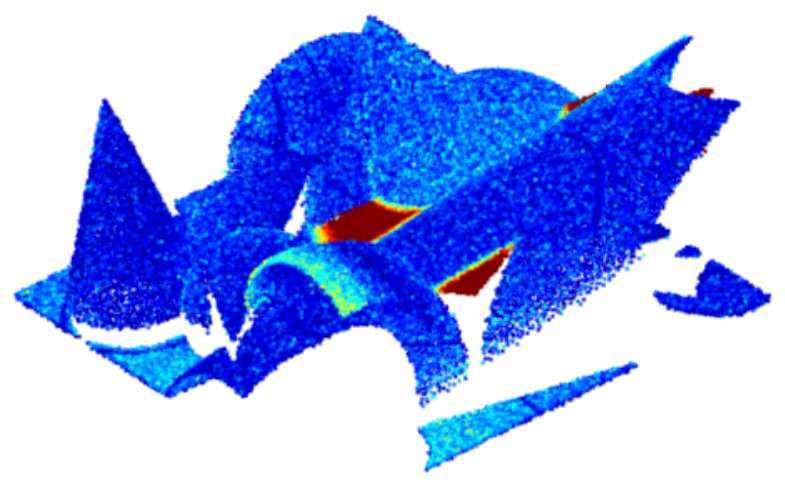}\hfill
	\includegraphics[width=.3\linewidth]{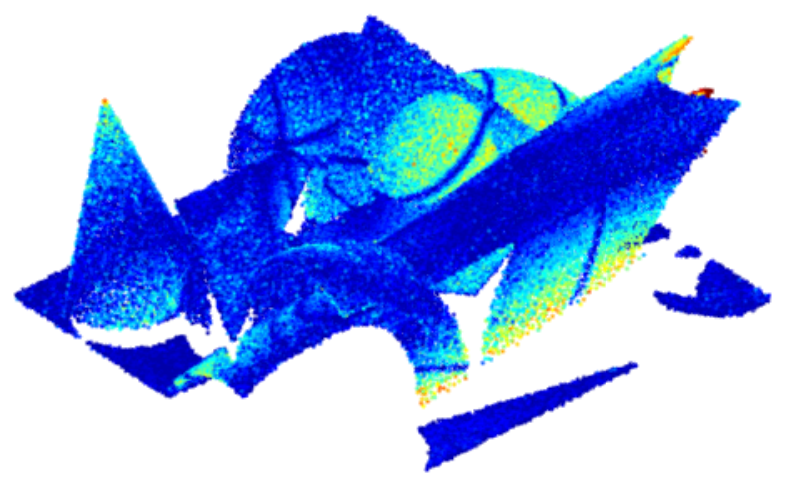}\hfill
	\includegraphics[width=.3\linewidth]{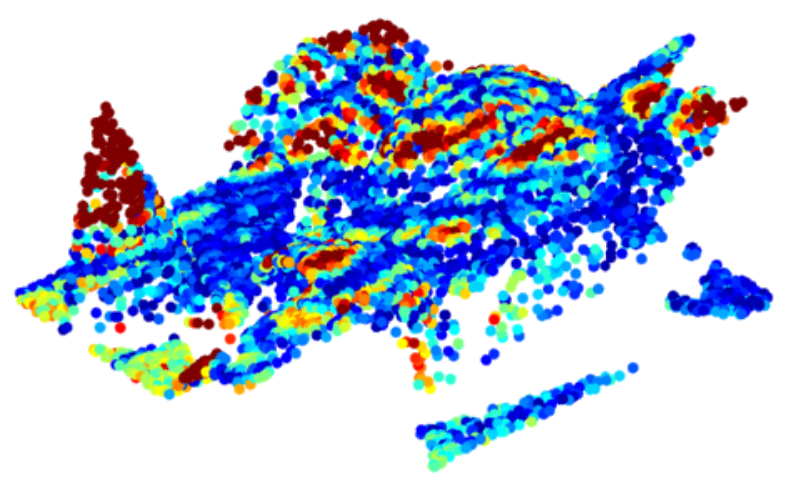}\vspace{1mm}
		\includegraphics[width=.14\linewidth]{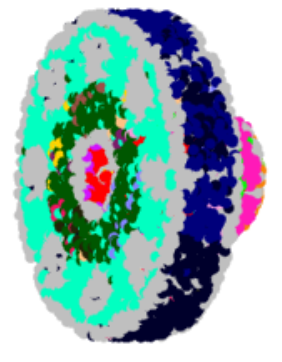}
		\includegraphics[width=.14\linewidth]{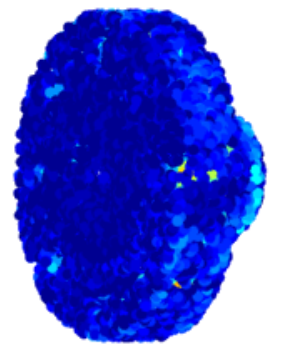}\hfill
		\includegraphics[width=.14\linewidth]{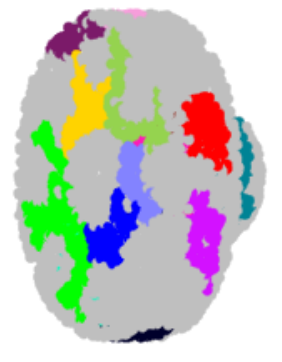}
		\includegraphics[width=.14\linewidth]{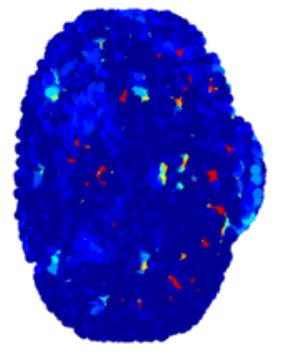}\hfill
		\includegraphics[width=.14\linewidth]{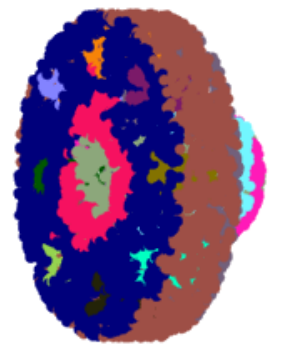}
		\includegraphics[width=.14\linewidth]{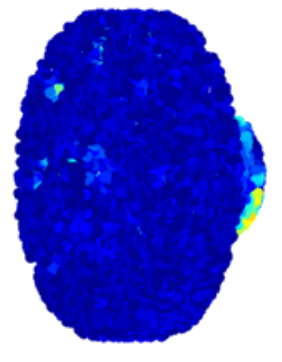}
	\caption{\textbf{(left)} Our method, \textbf{(middle)} Efficient RANSAC and \textbf{(right)} SPFN results on example point clouds from the \textit{primitect} (first two rows) and \textit{traceparts} (third row) datasets.
	We show detected instances (gray: unassigned) and distances from each point to its closest primitive (blue: 0, dark red: 0.1):
	our method has consistently smaller distances to the detected objects than Efficient RANSAC. 
	Being trained for object part detection, SPFN does not produce meaningful results on \textit{primitect}. 
	The concave cylinders on the \textit{traceparts} data are not detected by our method.
	}
	\label{fig:pointclouds}
\end{figure}

We report results on the \textit{traceparts} dataset in Table~\ref{table:traceparts} using their evaluation tool~\cite{li19}.
Note that our method is not well-suited for their data, since we only detect convex primitives, see Fig.~\ref{fig:pointclouds} (bottom).
On the other hand, running SPFN on \textit{primitect} data does not produce meaningful results at all, possibly because \textit{primitect} is too different from the \textit{traceparts} training data, see Fig.~\ref{fig:pointclouds} (top).

\begin{table}
\caption{Results on \textit{traceparts}: due to concave shapes, our scores are below those of SPFN for all reported metrics.
	Compared to Efficient RANSAC, we get a better primitive type accuracy.
	Note that SPFN without learning segmentation (SPFN-$\mathcal{L}_\textrm{seg}$) has a similar p-coverage to us. 
	}
\label{table:traceparts}
\begin{center}
	\renewcommand{\arraystretch}{1.25}
	\begin{tabular}{c c c c c}
		\toprule
		& SPFN-$\mathcal{L}_\textrm{seg}$ & SPFN & Eff. R. & ours \\
		\midrule
		p-coverage (0.01) & 62.2 & 88.3 & 65.7 & 63.8 \\
		p-coverage (0.02) & 77.7 & 96.3 & 88.6 & 78.3 \\
		primitive type (\%) & 92.4 & 96.9 & 52.9 & 77.3 \\
		\bottomrule
	\end{tabular}	
\end{center}
\end{table}

%

\subsection{Accuracy of detected objects}

Fig.~\ref{fig:coverages} (right) shows s-coverage on the \textit{primitect} data.
We outperform both the baseline and RANSAC by a large margin, which means that our parameter estimates are superior and fit the actual data better.

We also report the average DOD for all detected objects on \textit{primitect}, $\sigma=0.01$ in Table~\ref{table:segcomp_metrics}, with the same conclusion:
vote spreading combined with bin averaging and clustering-by-averaging largely improves the fit accuracy.
In particular, we are roughly twice as accurate as Efficient RANSAC.

\begin{figure}
	\includegraphics[width=\linewidth]{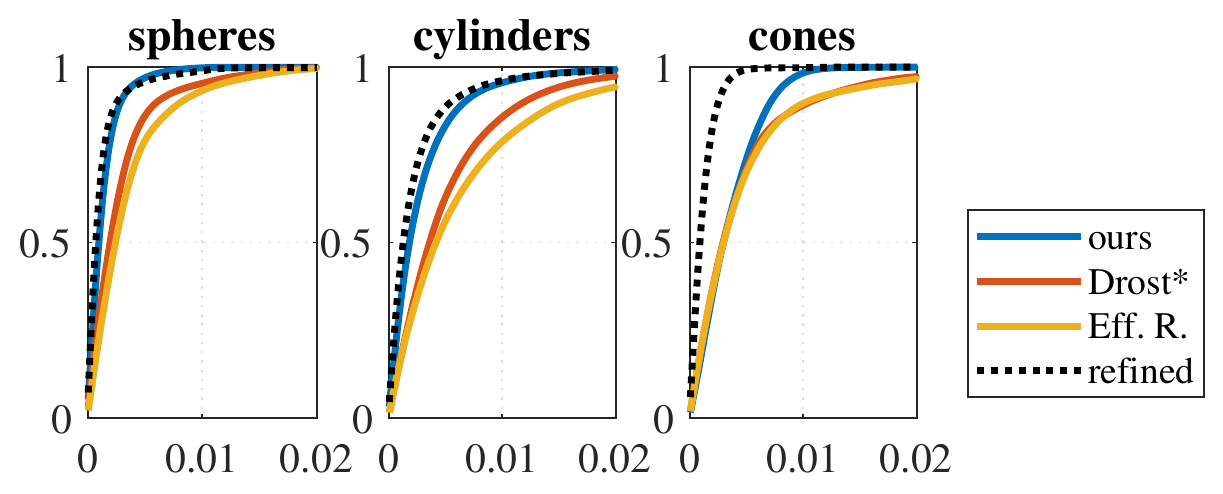}
	\caption{s-coverage for the \textit{primitive-rw} data: from left to right, we show average results for spheres, cylinders and cones. We clearly outperform both other detectors and for sphere and cylinder are even close to the result obtained via parameter optimization.}
	\label{fig:redwood_scov}
\end{figure}

To demonstrate testing on real data, we show the average s-coverage on \textit{primitive-rw}, separately for each primitive type. 
We applied robust refinement on the primitive parameters to get some notion of ``most-optimal parameter set,'' since ground truth is not available.
Note that such refinement can become computationally expensive if outlier ratio and noise level are unknown.
We show quantitative results in Fig.~\ref{fig:redwood_scov}, and visual output in Fig.~\ref{fig:redwood_results}.
Our method performs significantly better than both~\cite{drost15} and~\cite{schnabel07}.
For spheres, which have a 1D voting space, we even get close to the ``optimal'' results obtained by robust parameter fitting.

\subsection{Runtime}

Runtimes are summarized in Table~\ref{table:runtime}.
Our single-threaded implementation on an Intel Xeon CPU, is on par with the runtime of SPFN's GPU implementation on the Nvidia Titan X, and significantly faster than Efficient RANSAC (CPU).

\begin{table}
\caption{Runtime in ms: Our method, Efficient RANSAC~\cite{schnabel07,oesau19}, and SPFN~\cite{li19} on \textit{primitect}. We are significantly faster than the RANSAC on the full (160k) point cloud, and for cylinders and cones even outperform RANSAC on point clouds downsampled to 2048 points. Efficient RANSAC builds a tree as a pre-processing step, which takes an additional 220-240ms. SPFN's runtime on the GPU is similar to ours on the CPU.}
\label{table:runtime}
\begin{center}
	\renewcommand{\arraystretch}{1.25}
	\begin{tabular}{l c c c c c c} 
		\toprule
		&& joint & planes & spheres & cyl.s & cones\\
		\midrule
		Proposed Method & CPU & \textbf{79} & 23 & 36 & \textbf{38} & \textbf{58} \\
		Eff. R. (full) & CPU & 283 & 65 & 73 & 304 & 212 \\
		Eff. R. (2048 p.) & CPU & 226 & \textbf{7} & \textbf{14} & 150 & 93 \\
		SPFN (8096 p.) & GPU & 95 & - & - & - & - \\
		\bottomrule
	\end{tabular}	
\end{center}
\end{table}

\section{Discussion}
We have presented a novel method to group points into different primitives.
By deriving a new PPF-based local parameterization of cones and cylinders we can jointly assign one of four primitives to each point.
Experiments show that our method can outperform current state-of-the-art in terms of accuracy, robustness and speed.
Due to the low-dimensional voting space, our method can be run on low-power devices without the need for a GPU.


\addtolength{\textheight}{-7.5cm}   






\clearpage

\bibliographystyle{IEEEtran}
\bibliography{bibliography}

\begin{thebibliography}{10}
\providecommand{\url}[1]{#1}
\csname url@rmstyle\endcsname
\providecommand{\newblock}{\relax}
\providecommand{\bibinfo}[2]{#2}
\providecommand\BIBentrySTDinterwordspacing{\spaceskip=0pt\relax}
\providecommand\BIBentryALTinterwordstretchfactor{4}
\providecommand\BIBentryALTinterwordspacing{\spaceskip=\fontdimen2\font plus
\BIBentryALTinterwordstretchfactor\fontdimen3\font minus
  \fontdimen4\font\relax}
\providecommand\BIBforeignlanguage[2]{{%
\expandafter\ifx\csname l@#1\endcsname\relax
\typeout{** WARNING: IEEEtran.bst: No hyphenation pattern has been}%
\typeout{** loaded for the language `#1'. Using the pattern for}%
\typeout{** the default language instead.}%
\else
\language=\csname l@#1\endcsname
\fi
#2}}

\bibitem{drost10}
B.~Drost, M.~Ulrich, N.~Navab, and S.~Ilic, ``Model globally, match locally:
  Efficient and robust {3D} object recognition,'' in \emph{IEEE International
  Conference on Computer Vision and Pattern Recognition (CVPR)}, 2010.

\bibitem{drost15}
B.~Drost and S.~Ilic, ``Local {H}ough transform for {3D} primitive detection,''
  in \emph{International Conference on 3D Vision (3DV)}, 2015.

\bibitem{niblack90}
W.~Niblack and D.~Petkovic, ``On improving the acuracy of the {H}ough
  transform,'' \emph{Machine Vision and Applications}, vol.~3, 1990.

\bibitem{choi16}
S.~Choi, Q.~Zhou, S.~Miller, and V.~Koltun, ``A large dataset of object
  scans,'' \emph{arXiv:1602.02481}, 2016.

\bibitem{kaiser19}
A.~Kaiser, J.~A.~Y. Zepeda, and T.~Boubekeur, ``A survey of simple geometric
  primitives detection method for captured {3D} data,'' in \emph{Computer
  Graphics Forum}, 2019.

\bibitem{schnabel07}
R.~Schnabel, R.~Wahl, and R.~Klein, ``Efficient {RANSAC} for point-cloud shape
  detection,'' \emph{Computer Graphics Forum}, vol.~26, 2007.

\bibitem{li11}
Y.~Li, X.~Wu, Y.~Chrysathou, A.~Sharf, D.~Cohen-Or, and N.~J. Mitra,
  ``Glob{F}it: Consistently fitting primitives by discovering global
  relations,'' in \emph{ACM Transactions on Graphics (TOG)}, 2011.

\bibitem{kang15}
Z.~Kang and Z.~Li, ``Primitive fitting based on the efficient {multiBaySAC}
  algorithm,'' \emph{PloS one}, vol.~10, no.~3, 2015.

\bibitem{kim05}
S.~I. Kim and S.~J. Ahn, ``Extraction of geometric primitives from point cloud
  data,'' \emph{Energy}, vol.~2, 2005.

\bibitem{georgiev13}
K.~Georgiev and R.~Lakaemper, ``{RMSD}: A {3D} real-time mid-level scene
  description system,'' in \emph{IEEE Workshop on Robot Vision (WORV)}, 2013.

\bibitem{tran15}
T.-T. Tran, V.-T. Cao, and D.~Laurendeau, ``Extraction of reliable primitives
  from unorganized point clouds,'' \emph{3D Research}, vol.~6, no.~4, 2015.

\bibitem{sharma18}
G.~Sharma, R.~Goyal, D.~Liu, E.~Kalogerakis, and S.~Maji, ``{CSGNet}: Neural
  shape parser for constructive solid geometry,'' in \emph{IEEE International
  Conference on Computer Vision and Pattern Recognition (CVPR)}, 2018.

\bibitem{li19}
L.~Li, M.~Sung, A.~Dubrovina, L.~Yi, and L.~J. Guibas, ``Supervised fitting of
  geometric primitives to {3D} point clouds,'' in \emph{IEEE International
  Conference on Computer Vision and Pattern Recognition (CVPR)}, 2019.

\bibitem{hough62}
P.~V.~C. Hough, ``Method and means for recognizing complex patterns,'' 1962,
  {US} Patent 3,069,654.

\bibitem{duda72}
R.~O. Duda and P.~E. Hart, ``Use of the {Hough} transformation to detect lines
  and curves in pictures,'' \emph{Commun. ACM}, vol.~15, no.~1, 1972.

\bibitem{kiryati91}
N.~Kiryati and A.~M. Bruckstein, ``Antialiasing the hough transform,''
  \emph{CVGIP: Graphical Models and Image Processing}, vol.~53, 1991.

\bibitem{zhang96}
M.~Zhang, ``On the discretization of parameter domain in {Hough}
  transformation,'' in \emph{International Conference on Pattern Recognition},
  1996.

\bibitem{fernandes08}
L.~A. Fernandes and M.~M. Oliveira, ``Real-time line detection through and
  improved hough transform voting scheme,'' \emph{Pattern Recognition},
  vol.~41, 2008.

\bibitem{seib15}
V.~Seib, N.~Link, and D.~Paulus, ``Implicit shape models for {3D} shape
  classification with a continuous voting space,'' in \emph{International
  Conference on Computer Vision Theory and Applications (VISAPP)}, 2015.

\bibitem{ding05}
Y.~Ding, X.~Ping, M.~Hu, and D.~Wang, ``Range image segmentation based on
  randomized {H}ough transform,'' \emph{Pattern Recognition Letters}, vol.~26,
  no.~13, 2005.

\bibitem{tarshakurdi07}
F.~Tarsha-Kurdi, T.~Landes, and P.~Grussenmeyer, ``Hough-transform and extended
  {RANSAC} algorithms for automatic detection of {3D} building roof planes from
  {LIDAR} data,'' in \emph{ISPRS Workshop on Laser Scanning 2007 and SilviLaser
  2007}, vol.~36, 2007.

\bibitem{borrmann11}
D.~Borrmann, J.~Elseberg, K.~Lingemann, and A.~N{\"u}chter, ``The 3{D} {H}ough
  transform for plane detection in point clouds: A review and a new accumulator
  design,'' \emph{3D Research}, vol.~2, no.~2, p.~3, 2011.

\bibitem{limberger15}
F.~A. Limberger and M.~M. Oliveira, ``Real-time detection of planar regions in
  unorganized point clouds,'' \emph{Pattern Recognition}, vol.~48, 2015.

\bibitem{vera18}
E.~Vera, D.~Lucio, L.~A.~F. Fernandes, and L.~Velho, ``Hough transform for
  real-time plane detection in depth images,'' \emph{Pattern Recognition
  Letters}, vol. 103, 2018.

\bibitem{rabbani05}
T.~Rabbani and F.~V.~D. Heuvel, ``Efficient {Hough }transform for automatic
  detection of cylinders in point clouds,'' \emph{ISPRS WG III/3, III/4},
  vol.~3, 2005.

\bibitem{su10}
Y.-T. Su and J.~Bethel, ``Detection and robust estimation of cylinder features
  in point clouds,'' in \emph{ASPRS Conference}, 2010.

\bibitem{abuzaina13}
A.~Abuzaina, M.~S. Nixon, and J.~N. Carter, ``Sphere detection in {Kinect}
  point clouds via the {3D} {Hough} transform,'' in \emph{International
  Conference on Computer Analysis of Images and Patterns}, 2013.

\bibitem{figueiredo17}
R.~Figueiredo, P.~Moreno, and A.~Bernardino, ``Robust cylinder detection and
  pose estimation using {3D} point cloud information,'' in \emph{2017 IEEE
  International Conference on Autonomous Robot Systems and Competitions
  (ICARSC)}, 2017.

\bibitem{birdal18}
T.~Birdal, B.~Busam, N.~Navab, S.~Ilic, and P.~Sturm, ``A minimalist approach
  to type-agnostic detection of quadrics in point clouds,'' in \emph{IEEE
  International Conference on Computer Vision and Pattern Recognition (CVPR)},
  2018.

\bibitem{hodan18}
T.~Hodan, F.~Michel, E.~Brachmann, W.~Kehl, A.~G. Buch, D.~Kraft, B.~Drost,
  J.~Vidal, S.~Ihrke, X.~Zabulis, \emph{et~al.}, ``Bop: Benchmark for 6d object
  pose estimation,'' in \emph{European Conference on Computer Vision (ECCV)},
  2018.

\bibitem{bi03}
J.~J.~Bi and K.~P. Bennett, ``Regression error characteristic curves,'' in
  \emph{International Conference on Machine Learning (ICML)}, 2003.

\bibitem{hoover96}
A.~Hoover, G.~Jean-Baptiste, X.~Jiang, P.~J. Flynn, H.~Bunke, D.~B. Goldgof,
  K.~Bowyer, D.~W. Eggert, A.~Fitzgibbon, and R.~B. Fisher, ``An experimental
  comparison of range image segmentation algorithms,'' \emph{IEEE Transactions
  on Pattern Analysis and Machine Intelligence (PAMI)}, vol.~18, no.~7, 1996.

\bibitem{oesau19}
S.~Oesau, Y.~Verdie, C.~Jamin, P.~Alliez, F.~Lafarge, and S.~Giraudot, ``Point
  set shape detection,'' in \emph{{CGAL} User and Reference Manual},
  {4.14}~ed.\hskip 1em plus 0.5em minus 0.4em\relax {CGAL Editorial Board},
  2019.

\end{thebibliography}

\end{document}